\documentclass[twocolumn]{gewu}

\usepackage{wasysym}

\usepackage[toc,page,header]{appendix}

\usepackage{minitoc}
\usepackage{microtype}      
\usepackage[dvipsnames,table]{xcolor}         

\usepackage{hyperref}
\usepackage{url}
\usepackage{booktabs}
\usepackage{multirow}
\usepackage{subcaption}
\usepackage{caption}
\usepackage{amssymb}
\usepackage{graphicx}
\usepackage{amsmath}
\usepackage{wrapfig}
\usepackage{enumitem}
\usepackage{arydshln} 
\usepackage{gensymb}
\usepackage{makecell}
\usepackage{pifont} 
\usepackage{adjustbox}
\newsavebox{\tablebox}

\usepackage{tabularx}
\usepackage{booktabs}
\usepackage{colortbl}

\title{Crab$^{+}$: A Scalable and Unified Audio-Visual Scene Understanding Model with Explicit Cooperation}

\author[1,2,*]{Dongnuan~Cai}
\author[1,3,*]{Henghui~Du}
\author[3]{Chang~Zhou}
\author[3]{Xi~Chen}
\author[4]{Dan~Guo}
\author[5]{Hongyuan~Zhang}
\author[2]{Xuelong~Li}
\author[1,\textnormal{\Letter}]{Di~Hu}

\affiliation[1]{Gaoling School of Artificial Intelligence, Renmin University of China}
\affiliation[2]{Institute of Artificial Intelligence of China Telecom (TeleAI)}
\affiliation[3]{AI Technology Center, Online Video Business Unit, Tencent PCG}
\affiliation[4]{Hefei University of Technology}
\affiliation[5]{The University of Hong Kong}

\contribution[*]{Equal Contributions}
\contribution[\textnormal{\Letter}]{ Corresponding author}

\abstract{Developing Audio-Visual Large Language Models (AV-LLMs) for unified scene understanding is a pivotal pursuit in multimodal intelligence. While instruction tuning enables pre-trained models with multi-task abilities, we observe that conventional multi-task unification methods often suffer from severe negative transfer, where nearly 55\% of tasks degrade compared to single-task training. We attribute this phenomenon to audio-visual task heterogeneity, characterized by disparate task granularity and divergent capability demands, which lead to negative interference under joint training.
To tackle this, we present \textbf{Crab$^{+}$}, a scalable and unified audio-visual scene understanding model that addresses task heterogeneity through \textit{explicit cooperation} from both data and model perspectives.
From the \textit{data perspective}, we introduce AV-UIE v2, a comprehensive Audio-Visual Unified Instruction-tuning dataset with Explicit reasoning processes. It contains approximately 222K samples spanning 17 datasets and 7 tasks, enabling the model to uncover task-specific interactions and capture cross-task relationships at different levels of granularity. From the \textit{model perspective}, we design a unified input-output interface to align heterogeneous task formulations. To resolve divergent capability demands, we propose Interaction-aware LoRA (I-LoRA), which explicitly models inter-task relationships via dynamic routing to coordinate distinct audio-visual interaction patterns, thereby mitigating parameter interference.
Extensive experiments demonstrate that Crab$^{+}$ covers a broader range of tasks than existing unified models while outperforming specialized models on various benchmarks. We successfully reverse the negative transfer trend, achieving positive transfer where multi-task learning surpasses single-task baselines in nearly 88\% of tasks. These results are consistently observed across diverse AV-LLM paradigms and validated through in-depth visualization, suggesting Crab$^{+}$ serves as a robust step towards holistic audio-visual scene understanding.
}

\MyEmail{Dongnuan~Cai at \email{dongnuancai@ruc.edu.com}, Di~Hu at \email{dihu@ruc.edu.com}}
\checkdata[Code]{\url{https://github.com/GeWu-Lab/Crab_Plus}}

\begin{document}
\maketitle

\section{Introduction}
\label{intro}
Humans possess an inherent ability to use visual and auditory information to perform daily activities in a unified manner. For instance, humans can identify actions by integrating visual and auditory cues, localize events in space and time by fusing these modalities, and perform reasoning over complex audio-visual information to answer specific questions. From a cognitive science perspective, this unified capability facilitates transfer across related tasks, thereby promoting inter-task cooperation~\cite{taatgen2013nature}. Equipping machines with similar capabilities to process audio-visual data is thus crucial to the development of multimodal intelligence.

Recent advancements in foundation models have shifted the paradigm from task-specific systems to general-purpose architectures. In Natural Language Processing (NLP), Large Language Models (LLMs) unify diverse textual tasks under a single architectural framework. Inspired by the multimodal nature of the real world, Audio-Visual Large Language Models (AV-LLMs) extend the LLM framework to perform a wide range of complex audio-visual scene understanding tasks within a unified architecture. Building such models typically involves a two-stage training strategy: pre-training to align multimodal features with the LLM’s semantic space, followed by instruction tuning to equip the model with multi-task capabilities. In the second stage, most existing approaches~\cite{tang2025empowering,su2023pandagpt} apply direct instruction tuning to enable AV-LLMs to perform a variety of audio-visual tasks. However, given the inherent complexity of audio-visual tasks, it remains unclear whether such straightforward instruction tuning is sufficient to realize the potential of unified learning.

\begin{figure}[!t]
\centering
\includegraphics[width=0.48\textwidth]{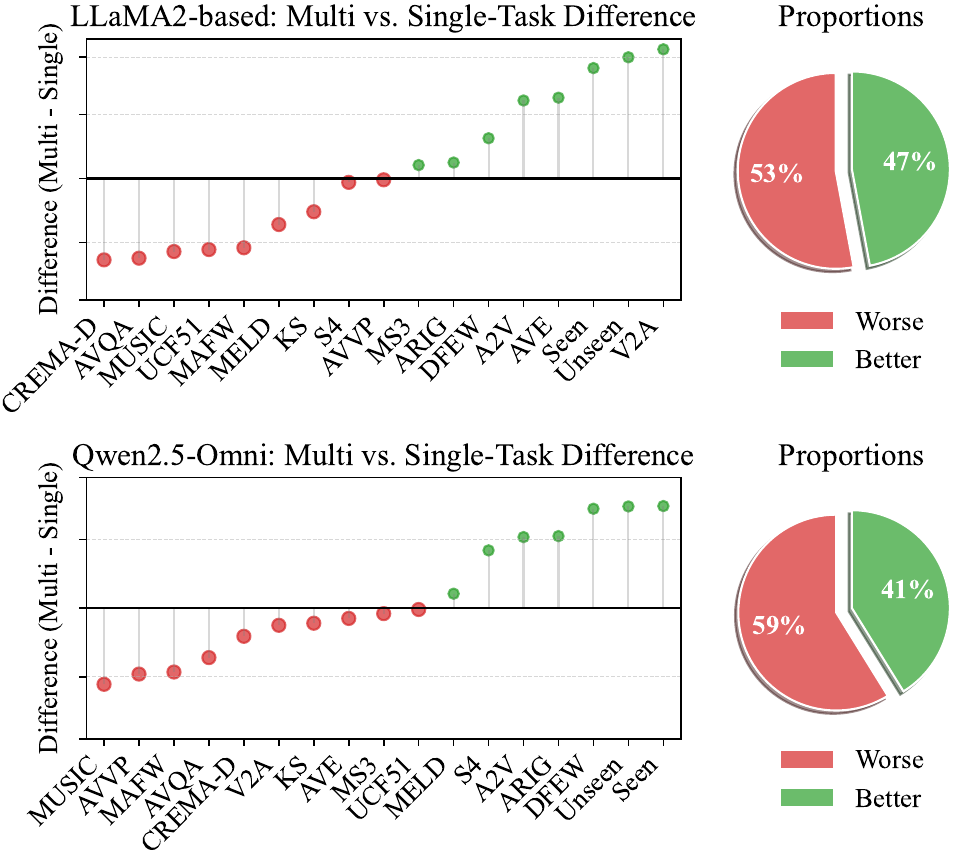}
\caption{Multi-task vs. single-task LoRA on customized LLaMA2 and Qwen2.5-Omni. Green and red cells highlight tasks with performance gains and drops, respectively, under multi-task learning.}
\label{obser}
\end{figure}

To investigate this, we follow prior work~\cite{tang2025empowering,su2023pandagpt} and benchmark two representative AV-LLMs (a LLaMA2-based model~\cite{touvron2023llama} and Qwen2.5-Omni~\cite{xu2025qwen2}) via LoRA fine-tuning on standard instruction–response supervision, which consists of independently annotated instruction–output pairs, covering 7 diverse audio-visual tasks across 17 datasets. While multi-task learning aims to leverage inter-task synergies, our empirical results reveal a counter-intuitive phenomenon: naïve task unification performs worse than single-task training in nearly 56\% of the evaluated settings (Figure~\ref{obser}). We attribute this performance degradation to two fundamental bottlenecks arising from inherent task heterogeneity. \textit{First}, disparate granularity of audio-visual tasks hinders effective cooperation under unified learning. These tasks operate at different levels of task granularity. For example, low-level grounding tasks involve aligning and modeling audio-visual information across temporal and spatial dimensions. Conversely, high-level reasoning tasks require a causal understanding of semantics, affect, and action based on spatio-temporal inputs. Direct instruction tuning fails to bridge these differences without intermediate representations. Consequently, this gap impedes effective cooperation between tasks. \textit{Second}, divergent audio-visual capability demands may lead to parameter interference during optimization. Different tasks impose distinct requirements on the models. Temporal localization tasks demand perception and alignment capabilities over time. Spatial localization tasks necessitate modeling the correspondence between audio signals and visual information at both the region and pixel levels. Furthermore, spatio-temporal reasoning tasks require the model to handle semantic reasoning within complex contexts. Existing parameter-efficient adaptation methods~\cite{hu2022lora} typically rely on static and shared adaptation schemes, which lack the flexibility to accommodate such divergent interaction patterns, often resulting in negative transfer under joint optimization.

\begin{figure*}[!t]
\centering
\includegraphics[width=0.95\textwidth]{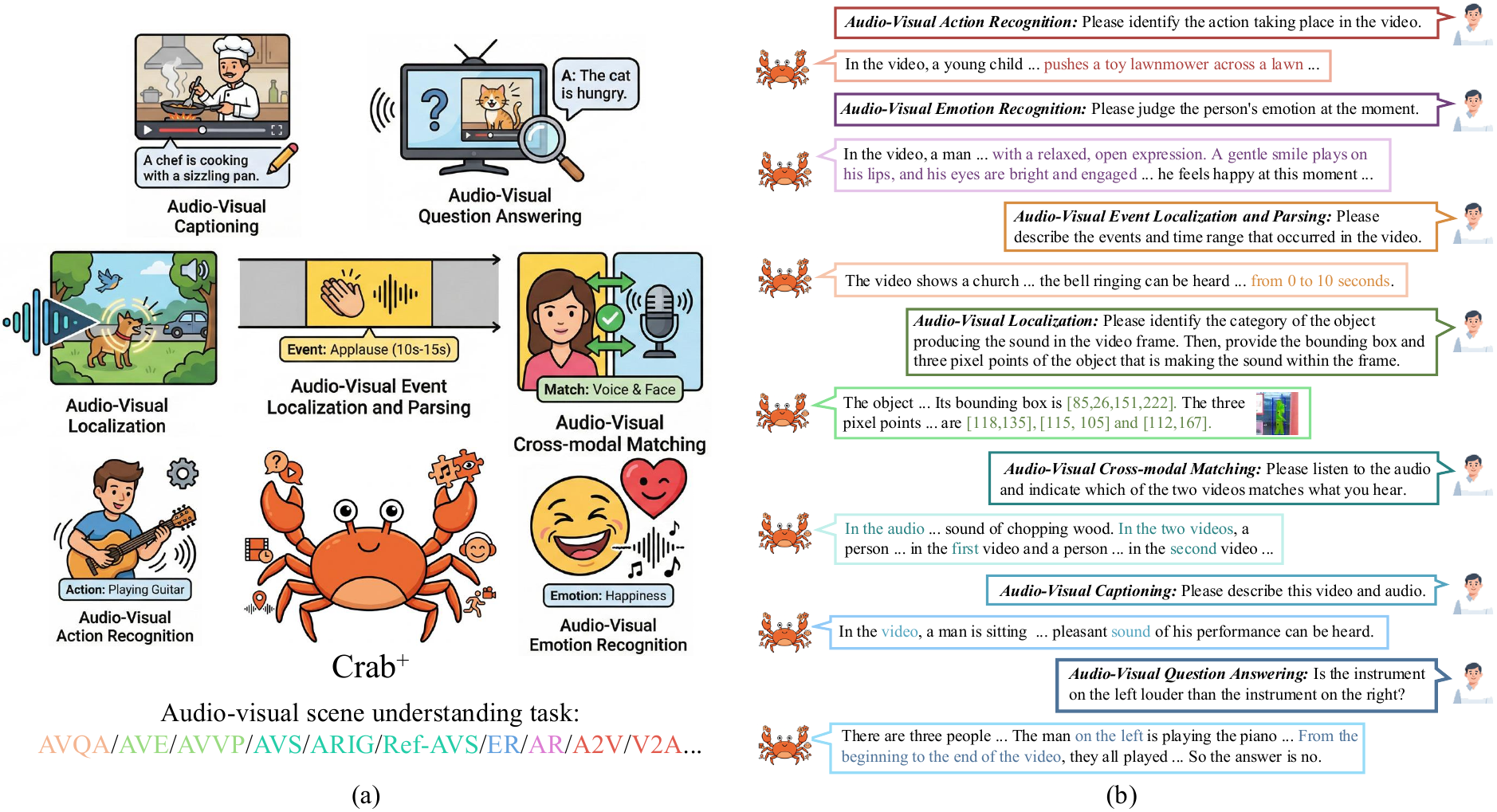}
\caption{Overview of Crab$^{+}$. (a) Scalable and unified architecture capable of addressing diverse audio-visual scene understanding tasks. (b) Explicit cooperation that captures complex cross-task dependencies to improve overall performance.}
\label{teaser}
\end{figure*}

To address these limitations, we introduce \textbf{Crab$^{+}$}, a scalable and unified audio-visual scene understanding model that tackles task heterogeneity through explicit cooperation from both data and model perspectives. From the \textbf{data perspective}, we construct AV-UIE v2, a large-scale \textbf{A}udio-\textbf{V}isual \textbf{U}nified \textbf{I}nstruction tuning dataset with \textbf{E}xplicit reasoning processes, scaled up to approximately 222K samples spanning 7 tasks and 17 datasets. Instead of naïvely aggregating data, we use explicit reasoning processes as an intermediate supervision-level representation to reduce semantic inconsistency arising from differences in task granularity. From the \textbf{model perspective}, we develop a unified input-output interface by formulating all task targets as sequences, mapping diverse task-specific outputs into a unified sequence representation. To alleviate parameter interference induced by divergent capability demands, we further propose Interaction-aware LoRA (I-LoRA). This module extends conventional static adaptation by employing a dynamic routing mechanism to explicitly model complex inter-task relationships. By using a learnable router to adaptively route input tokens to appropriate LoRA heads, I-LoRA decouples conflicting audio-visual interaction patterns, thereby satisfying distinct task requirements and enabling effective unified learning. As illustrated in Fig.~\ref{teaser}, Crab$^{+}$ supports the execution of diverse audio-visual scene understanding tasks within a single unified model, while effectively captures the inter-task dependencies to foster cooperation.

Overall, this work substantially extends our preliminary conference version~\cite{du2025crab} in terms of data construction, model architecture, and experiment analysis. From a data perspective, we have significantly expanded the coverage of audio-visual scene understanding by upgrading AV-UIE to AV-UIE v2. This upgrade scales the number of source datasets from 8 to 17 and broadens the scope of tasks from 4 to 7, incorporating new tasks such as audio-visual emotion recognition, action recognition, and cross-modal matching. These additions allow the model to accommodate a wider spectrum of complex audio-visual scenarios. From the model perspective, we refine the unified input-output interface to incorporate pixel-level perception alongside semantic understanding. This design obviates the need for disjoint optimization of separate modules, establishing a streamlined and unified single-stage training pipeline. In terms of experiment analysis, we not only benchmark the performance of Crab$^{+}$ but also apply our method across three representative AV-LLM construction paradigms to assess its generalizability: LLMs equipped with visual and audio branches (LLM+V+A), Vision-LLMs enriched with audio capabilities (V-LLM+A), and native AV-LLMs. Extensive visualization and empirical analysis further substantiate the effectiveness of our approach in fostering inter-task synergy. Our main contributions are summarized as follows:\looseness=-1
\begin{itemize}
\item We propose Crab$^{+}$, a scalable and unified audio-visual scene understanding model that tackles task heterogeneity through explicit cooperation, thereby facilitating positive transfer across heterogeneous~tasks.\looseness=-1
\item We introduce AV-UIE v2, a large-scale dataset comprising nearly 222K samples spanning 7 tasks and 17 datasets. It uses explicit reasoning processes as an intermediate supervision-level representation to reduce semantic inconsistency arising from differences in task granularity.\looseness=-1
\item We design a unified input-output interface to standardize task execution, coupled with Interaction-aware LoRA (I-LoRA) to alleviate parameter interference. Using dynamic routing, I-LoRA decouples conflicting audio-visual interaction patterns to satisfy distinct task requirements.
\end{itemize}

\section{Related Works} 
\subsection{Audio-Visual Scene Understanding}
\label{sec:av_learning}

Audio-visual scene understanding achieves a comprehensive perception of the environment by integrating auditory and visual modalities~\cite{feng2025towards,wang2025erf,wei2022learning}. This field encompasses diverse tasks, including basic recognition such as audio-visual action recognition~\cite{gao2020listen}, fine-grained perception including temporal localization~\cite{tian2018audio,tian2020unified} and segmentation~\cite{zhou2022audio,wang2024ref}, as well as complex reasoning like audio-visual question answering~\cite{yang2022avqa,li2022learning}.
Traditional research has predominantly relied on task-specific architectures. Specialized modules, such as audio-guided attention mechanisms~\cite{tian2018audio} and hierarchical spatio-temporal networks~\cite{li2022learning}, are designed to model cross-modal correlations for task-specific objectives. While effective within specific task boundaries, these models exhibit limited generalization across tasks. Recently, the emergence of AV-LLMs~\cite{zhang2023video,su2023pandagpt} has marked a significant transition from specialized experts to unified generalists. By leveraging the rich semantic priors of LLMs and aligning multimodal features into a unified semantic space, these models enable more general-purpose audio-visual understanding within a single framework. However, equipping AV-LLMs with multi-task capabilities via instruction tuning introduces significant challenges, particularly the bottleneck of negative transfer stemming from audio-visual task heterogeneity. Distinct from existing methods that overlook such interference, our work investigates the root causes of task heterogeneity and mitigates this bottleneck through explicit cooperation between data and model perspectives.

\subsection{Unified Learning Paradigm}
The pursuit of Artificial General Intelligence (AGI) has driven a shift in machine learning from specialized models toward unified frameworks. Rather than tailoring architectures to specific tasks, recent research increasingly emphasizes foundation models with broad generalization capabilities. This section reviews the evolution of this paradigm from the perspectives of architectural unification and multimodal integration.

\textbf{Unified Architecture}. Historically, neural architectures were designed with strong inductive biases tailored to specific data structures: CNNs for grid-structured data and RNNs for sequential dependencies. The emergence of the Transformer~\cite{vaswani2017attention} marked a shift toward architectural unification by introducing a shared attention-based mechanism. By representing diverse inputs as token sequences, this architecture largely decouples network design from task-specific constraints. Building on this, recent advancements have adopted this framework, often initialized from LLMs, as a general-purpose backbone~\cite{grattafiori2024llama}. This universality facilitates a scalable approach to general-purpose modeling, allowing a single architecture to adapt to diverse downstream tasks.

\textbf{Unified Modality}. Parallel to architectural convergence, machine perception has evolved from isolated unimodal analysis toward integrated multimodal understanding. A pivotal advancement occurred with vision-language pre-training, where contrastive frameworks like CLIP~\cite{radford2021learning} and ALIGN~\cite{jia2021scaling} aligned visual and textual representations into a shared semantic space. Building on these foundations, Multimodal Large Language Models (MLLMs)~\cite{li2023blip,liu2023visual} have further integrated visual information into LLMs, equipping models with comprehensive multimodal understanding capabilities. Current research is extending beyond the vision-language boundary to encompass auditory signals. Through the simultaneous integration of audio, visual, and linguistic modalities, these approaches endeavor to construct unified interfaces capable of seamlessly perceiving and analyzing complex environments.\looseness=-1

While the unified learning paradigm provides a promising foundation for general-purpose intelligence, its application to audio-visual scenarios is challenged by negative interference induced by task heterogeneity. Motivated by this, we focus on mitigating such interference within a unified framework by explicitly addressing heterogeneity at both the data and model levels, enabling more robust audio-visual scene understanding.

\section{The AV-UIE v2 Dataset} 

As analyzed in Section~\ref{intro}, diverse audio-visual tasks vary significantly in granularity. Existing audio-visual datasets mostly lack a suitable intermediate representation to bridge these tasks, which limits their ability to facilitate inter-task synergy under unified learning effectively. To alleviate this, we introduce AV-UIE v2, an Audio-Visual Unified Instruction-tuning dataset with Explicit reasoning processes. AV-UIE is an instruction tuning dataset covering diverse audio-visual scene understanding tasks and scenarios. By utilizing explicit reasoning processes as intermediate representations, AV-UIE establishes reasonable connections across different tasks, thereby facilitating unified learning. The preliminary version of this dataset was introduced in our conference paper~\cite{du2025crab}. In this work, we present AV-UIE v2, which significantly expands both the scale and diversity of the dataset by increasing the number of source datasets from 8 to 17 and extending the task scope from 4 to 7. New tasks include emotion recognition, action recognition, and a cross-modal matching subset derived from VGGSound~\cite{chen2020vggsound}. We detail the overall construction pipeline and provide statistical analysis in the following sections.

\begin{figure*}[!t]
\centering
\includegraphics[width=0.95\textwidth]{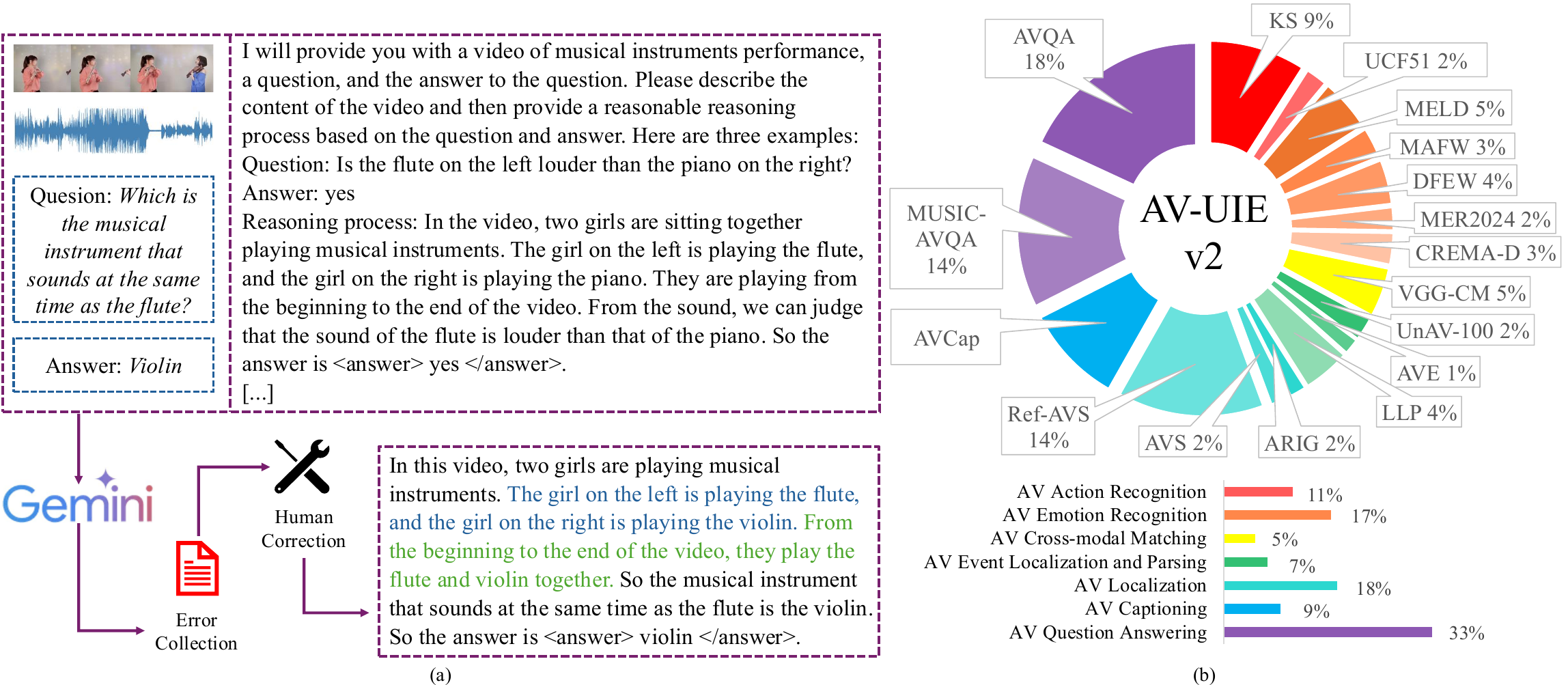}
\caption{Overview of AV-UIE v2 dataset.  
(a) Construction pipeline. We leverage MLLMs to augment raw data and simple labels with explicit reasoning processes. (b) Statistical distribution. The dataset integrates a comprehensive collection of source data from 17 datasets, covering 7 distinct scene understanding tasks.}
\label{dataset}
\end{figure*}

\begin{table}[t!]
\caption{Data composition of the AV-UIE v2 dataset, organized by task complexity from basic recognition to complex reasoning.}
\label{tab:av_datasets}
\centering
\footnotesize
\renewcommand{\arraystretch}{1}
\setlength{\tabcolsep}{5pt}

\resizebox{\columnwidth}{!}{
\begin{tabular}{l l c r}
\toprule
\textbf{Audio-Visual Task} & \textbf{Source Dataset} & \textbf{Domain} & \textbf{Samples (K)} \\
\midrule
\multirow{2}{*}{Action Recognition} 
 & Kinetics-Sounds~\cite{arandjelovic2017look} & Human Actions & 20.1 \\
 & UCF51~\cite{soomro2012ucf101} & Human Actions & 4.5 \\
\midrule
\multirow{5}{*}{Emotion Recognition} 
 & MELD~\cite{poria2018meld} & Conversations & 11.0 \\
 & MAFW~\cite{liu2022mafw} & Movies \& TV & 6.4 \\
 & DFEW~\cite{jiang2020dfew} & Movies \& TV & 9.3 \\
 & MER2024~\cite{lian2024mer} & Movies \& TV & 4.8 \\
 & CREMA-D~\cite{cao2014crema} & Studio Speech & 6.6 \\
\midrule
Cross-modal Matching & VGG-CM~\cite{chen2020vggsound} & Open-domain & 10.3 \\
\midrule
\multirow{3}{*}{Event Localization and Parsing}
 & UnAV-100~\cite{geng2023dense} & Open-domain & 4.9 \\
 & AVE~\cite{tian2018audio} & Open-domain & 3.3 \\
 & LLP~\cite{tian2020unified} & Open-domain & 9.9 \\
\midrule
\multirow{3}{*}{Localization} 
 & ARIG~\cite{zhou2022audio} & Open-domain & 3.4 \\
 & AVS~\cite{zhou2022audio} & Open-domain & 4.0 \\
 & Ref-AVS~\cite{wang2024ref} & Open-domain & 30.7 \\
\midrule
Captioning & AVCap~\cite{chen2023valor} & Open-domain & 20.7 \\
\midrule
\multirow{2}{*}{Question Answering} 
 & MUSIC-AVQA~\cite{li2022learning} & Musical & 31.9 \\
 & AVQA~\cite{yang2022avqa} & Open-domain & 40.2 \\
\bottomrule
\end{tabular}
}
\end{table}

\subsection{Dataset Construction}
We construct our dataset by integrating diverse existing audio-visual sources. To address the inconsistent task granularity across these datasets, we convert the original annotations into detailed textual descriptions containing explicit reasoning processes. This serves as a unified representation to establish connections between diverse tasks. As shown in Figure~\ref{dataset} (a), each data sample combines the raw video and audio, the ground-truth label, and a task-specific prompt. We utilize Gemini 1.5 Pro~\cite{team2024gemini} via in-context learning to generate the reasoning processes. To ensure data quality, we implement a two-step verification process. First, we require the generated reasoning to logically conclude with the original ground-truth label. Samples that fail this consistency check are automatically discarded. Second, we perform a manual human review on a random subset of the valid samples. For instance, as shown in Figure~\ref{dataset} (a), a sample simply labeled ``violin'' (answering the question ``Which musical instrument sounds at the same time as the flute?'') is expanded into a detailed sequence describing the acoustic characteristics and temporal synchronization of both instruments.

We follow this pipeline to process multiple existing datasets. These include MUSIC-AVQA~\cite{li2022learning}, AVQA~\cite{yang2022avqa}, AVE~\cite{tian2018audio}, UnAV-100~\cite{geng2023dense}, LLP~\cite{tian2020unified}, MELD~\cite{poria2018meld}, CREMA-D~\cite{cao2014crema}, MAFW~\cite{liu2022mafw}, DFEW~\cite{jiang2020dfew}, MER2024~\cite{lian2024mer}, Kinetics-Sounds (KS)~\cite{arandjelovic2017look,kay2017kinetics}, UCF51~\cite{soomro2012ucf101}, and our proposed VGGSound Cross-modal Matching (VGG-CM) dataset. To address the lack of specific temporal intervals in the original LLP training data, we leverage the temporal reasoning capabilities of MLLMs to generate interval annotations. For AVS-Bench~\cite{zhou2022audio} and Ref-AVS~\cite{wang2024ref}, we format the output to support spatial localization and segmentation tasks. For localization, we extract the bounding box from the object mask as $[x_{left}, y_{top}, x_{right}, y_{bottom}]$. For segmentation, we build upon this by additionally sampling three points within the mask via preprocessing to guide pixel-level localization. Furthermore, we incorporate the audio-visual perception dataset proposed by VALOR~\cite{chen2023valor} to enrich our formats. This process creates the AV-UIE v2 dataset under a rigorous quality control protocol.\looseness=-1

\subsection{Dataset Analysis}

The AV-UIE v2 dataset contains approximately 222K samples, each annotated with explicit reasoning processes to establish connections between diverse tasks. As detailed in Table \ref{tab:av_datasets} and visualized in Figure \ref{dataset} (b), the data distribution follows a hierarchical structure designed to evaluate the model's capability on tasks of varying granularity. 
At the foundational level, basic recognition tasks constitute a substantial portion of the dataset, including action recognition (11\%) and emotion recognition (17\%). These tasks aim to identify human activities and affective states from multimodal inputs (e.g., Kinetics-Sounds and MELD).
Building on this foundation, the dataset incorporates fine-grained spatio-temporal perception tasks that require precise cross-modal alignment. This includes cross-modal matching (5\%), event localization and parsing (7\%), and localization (18\%). In these settings, models are required to ground audio cues within corresponding spatial or temporal visual contexts (e.g., Ref-AVS).
At the highest level, complex reasoning tasks account for the largest share of the dataset. This level comprises captioning (9\%) and question answering (33\%). Such tasks demand the integration of multimodal information to generate semantic descriptions or answer challenging queries (e.g., MUSIC-AVQA and AVQA).

\section{Method}

\begin{figure}[!t]
\centering
\includegraphics[width=0.50\textwidth]{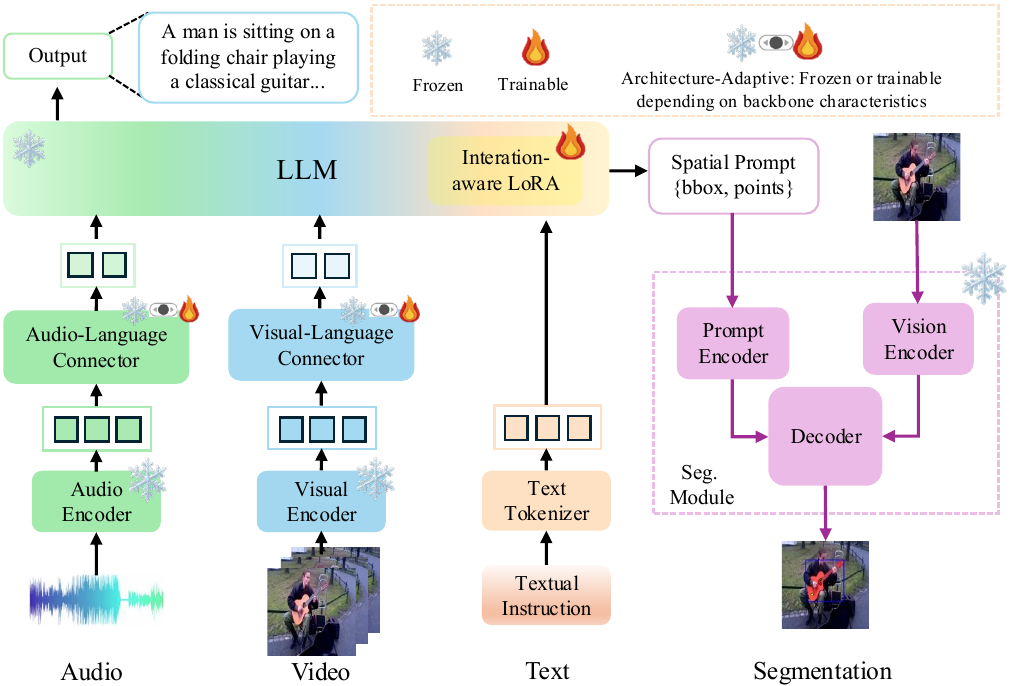}
\caption{Our unified input-output interface comprises a visual branch, an audio branch, a LLM backbone, and a segmentation module, trained on the AV-UIE v2 dataset with interaction-aware LoRA.\looseness=-1}
\label{model}
\end{figure}

To address the limitations discussed in Section~\ref{intro}, we present a unified model that standardizes task formulations via a consistent input-output interface, integrating Interaction-aware LoRA (I-LoRA) to mitigate inter-task interference. This methodology is designed to be architecture-agnostic. We primarily implement our method on the native AV-LLM paradigm (Crab$^{+}$) to test the performance upper bound, benefiting from its joint pre-training for strong cross-modal alignment. To further demonstrate the generalizability of our approach, we also extend it to composite paradigms (i.e., LLM+V+A and V-LLM+A). The following subsections detail the unified interface, architectural instantiations, and the I-LoRA mechanism.\looseness=-1

\subsection{Unified Input-output Interface}
\label{interface}
To bridge the gap between diverse levels of task granularity, we design a unified interface to cast all task targets as sequences, as shown in Figure~\ref{model}, enabling a streamlined single-stage training pipeline and avoiding disjoint optimization.\looseness=-1

\textbf{Visual Branch.} The visual branch consists of a pre-trained visual encoder $\mathcal{E}_{V}(\cdot)$ and a connector $\psi_V(\cdot)$, which projects visual features into the LLM's embedding space. Given a sequence of $T_v$ video frames $V = \{ I_i \in \mathbb{R}^{H \times W \times C} \}_{i=1}^{T_v}$, where $H$, $W$, and $C$ denote the height, width, and number of channels, respectively, each frame is partitioned into $N_v = \frac{H \times W}{P^2}$ non-overlapping patches. These patches are linearly mapped into a $D_v$-dimensional embedding space, yielding patch-level features $f_v^i \in \mathbb{R}^{N_v \times D_v}$ for the $i$-th frame. The aggregated visual representation is denoted as $F_v = \{ f_v^i \}_{i=1}^{T_v}$. The connector $\psi_V(\cdot)$ then transforms and flattens these features to match the input dimension of the LLM, producing the final visual tokens $H_v \in \mathbb{R}^{L_v \times h}$, where $L_v$ is the number of visual tokens and $h$ is the hidden size of the LLM.\looseness=-1

\textbf{Audio Branch.} Analogously, the audio branch employs a pre-trained audio encoder $\mathcal{E}_{A}(\cdot)$ and a connector $\psi_A(\cdot)$ to align audio features with the language embeddings. Given a sequence of $T_a$ audio segments $A = \{ S_i \in \mathbb{R}^{L_s} \}_{i=1}^{T_a}$, where $L_s$ denotes the length of each segment, each segment is mapped into a $D_a$-dimensional embedding space, yielding segment-level features $f_a^i \in \mathbb{R}^{N_a \times D_a}$ for the $i$-th segment. The aggregated audio representation is denoted as $F_a = \{ f_a^i \}_{i=1}^{T_a}$. The connector $\psi_A(\cdot)$ then transforms and flattens these features to match the input dimension of the LLM, producing the final audio tokens $H_a \in \mathbb{R}^{L_a \times h}$, where $L_a$ is the number of audio tokens and $h$ is the hidden size of the LLM.

\textbf{Token Fusion and LLM Backbone.} 
Following feature encoding, the multimodal tokens are concatenated with text tokens derived from the tokenizer. This unified sequence is fed into the LLM backbone for joint autoregressive generation, enabling integrated reasoning across audio, visual, and textual modalities within a single language modeling interface.

\textbf{Segmentation Module.}
Previous approaches\cite{lai2024lisa,ren2024pixellm} typically incorporate segmentation into MLLMs via special tokens, often necessitating multi-stage training. Inspired by Seg-Zero\cite{liu2025seg}, we adopt a decoupled reasoning-segmentation method by leveraging SAM2\cite{ravi2024sam} as a frozen segmentation module, as illustrated in Figure~\ref{model}. The MLLM predicts spatial prompts (e.g., bounding boxes and points), which are then used to query SAM2 for mask generation. This design effectively decouples multimodal grounding from segmentation, facilitating a unified single-stage architecture.
Formally, given a training frame $I \in \mathbb{R}^{H \times W \times 3}$ and its ground-truth binary mask $M \in \{0,1\}^{H \times W}$, we formulate the supervision target $\mathcal{Y}_{\text{seg}}$. This target comprises the minimal enclosing bounding box $b$ and a set of points derived via a geometry-based sampling strategy. We select the centers of the three largest inscribed circles within the mask region, allowing partial overlaps among the circles but enforcing a maximum Intersection over Union (IoU) constraint between any pairs. These centers serve as point prompts $\{p_1, p_2, p_3\}$, yielding the supervision set:
\begin{equation}
\mathcal{Y}_{\text{seg}} = \{ b, p_1, p_2, p_3 \}.
\end{equation}
During inference, given the image $I$, audio $A$, and instruction $X_{\text{text}}$, the model predicts these spatial parameters~directly:\looseness=-1
\begin{equation}
(\hat{b}, \hat{p}_1, \hat{p}_2, \hat{p}_3)
= \mathcal{F}_{\text{MLLM}}(I, A, X_{\text{text}}),
\end{equation}
where $\mathcal{F}_{\text{MLLM}}(\cdot)$ denotes the multimodal large language model. The predicted spatial prompts are then encoded by SAM2's prompt encoder $\mathcal{E}_{\text{prompt}}(\cdot)$ and combined with the image features from SAM2's vision encoder $\mathcal{E}_{\text{vision}}(\cdot)$ via the mask decoder $\mathcal{D}_{\text{mask}}(\cdot)$ to produce the final mask:
\begin{equation}
\hat{M} = \mathcal{D}_{\text{mask}}\big(
\mathcal{E}_{\text{prompt}}(\hat{b}, \hat{p}_1, \hat{p}_2, \hat{p}_3),
\mathcal{E}_{\text{vision}}(I)
\big).
\end{equation}

Building on the aforementioned branches and modules, our approach establishes a unified input-output interface that accepts audio-visual inputs alongside task-specific instructions, and generates corresponding task-oriented~outputs.\looseness=-1

\subsection{Architectural Instantiations}

\textbf{The Native AV-LLM Implementation.}
The primary instantiation of Crab$^{+}$ adopts the native AV-LLM paradigm, pre-trained on large-scale audio-visual data. We build upon Qwen2.5-Omni~\cite{xu2025qwen2}. The visual branch utilizes the Qwen2.5-VL~\cite{bai2025qwen2} backbone, a vision transformer with window attention for efficient spatial modeling, followed by a merger for token compression. The audio branch follows the Qwen2-Audio~\cite{chu2024qwen2} architecture, employing a Whisper-Large-v3~\cite{radford2023robust} encoder connected via a linear projection layer.

\textbf{Extension to Composite Architectures.}
To evaluate the robustness of our method across diverse architectures, we apply it to composite paradigms where modalities are bridged via lightweight connectors. For the LLM+V+A setting, we utilize LLaMA2-7B-Chat~\cite{touvron2023llama} and LLaMA3-8B-Instruct~\cite{grattafiori2024llama} as text backbones, integrating independent visual (CLIP-ViT-L/14~\cite{radford2021learning}) and audio (BEATs~\cite{chen2022beats}) encoders. Each modality is aligned via a Q-Former~\cite{li2023blip} followed by a two-layer MLP. This setup tests the method's ability to coordinate loosely coupled modalities. Conversely, the V-LLM+A setting augments an existing vision-language model, Qwen2.5-VL~\cite{bai2025qwen2}, with an additional audio branch consisting of BEATs, a window-level Q-Former~\cite{tang2023salmonn}, and a subsequent two-layer MLP. A key difference lies in initialization: unlike the native model, which benefits from checkpoints with prior cross-modal alignment, these composite architectures combine independent encoders. Thus, they require an initial connector alignment stage before instruction tuning. By validating our approach on these variants alongside the native Crab$^{+}$, we ensure the method is robust across different model construction paradigms.

\subsection{Interaction-aware LoRA}

\begin{figure}[!t]
\centering
\includegraphics[width=0.50\textwidth]{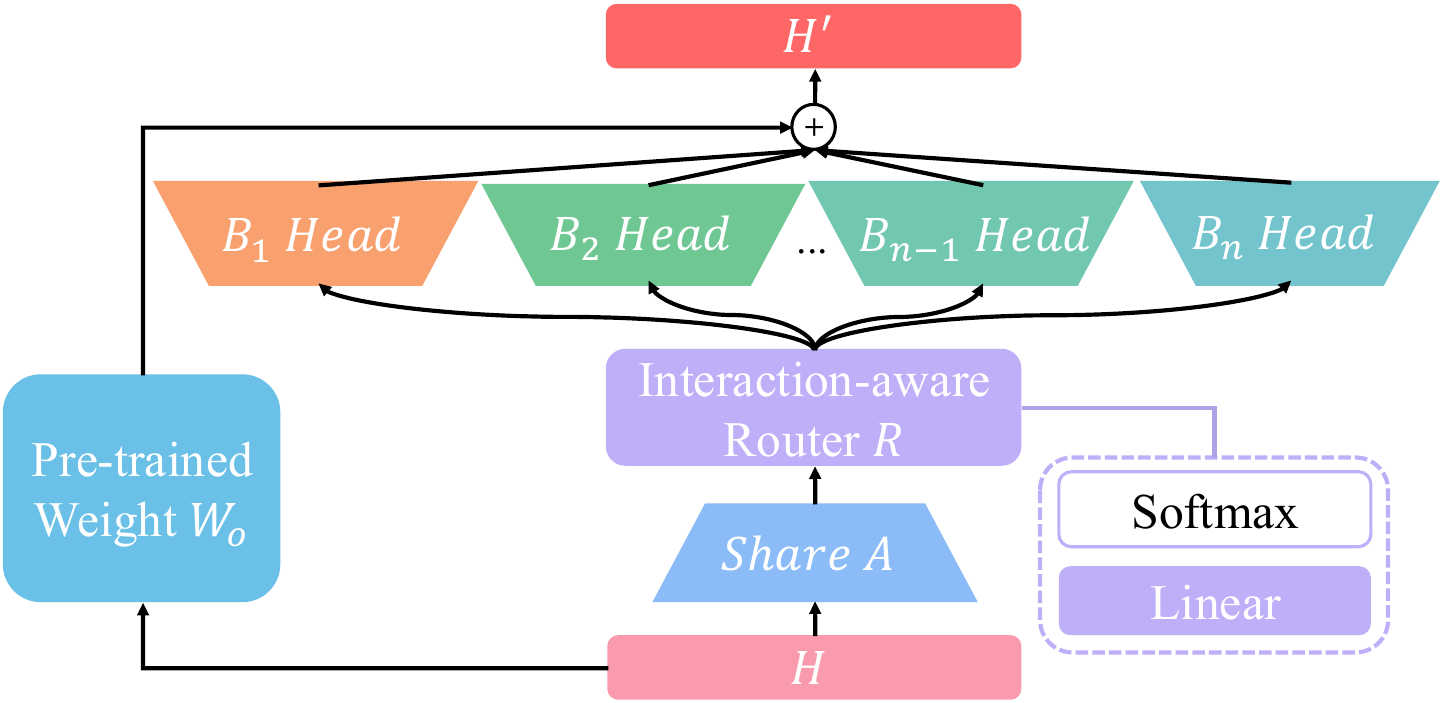}
\caption{Interaction-aware LoRA comprises a shared matrix $A$ and multiple $B$ matrices, where inputs are dynamically routed to specific $B$ matrices.}
\label{lora}
\end{figure}

LoRA~\cite{hu2022lora} achieves parameter efficiency by decomposing weight updates into low-rank matrices. However, in multi-task learning, enforcing shared static parameters across diverse tasks limits flexibility. Heterogeneous task demands often induce parameter interference, resulting in negative transfer. While Multi-LoRA~\cite{wang2023multilora} addresses this by assigning disjoint adapters to individual tasks, such parameter isolation prevents the model from exploiting the shared representations inherent in audio-visual data. To overcome this, we propose Interaction-aware LoRA (I-LoRA), a dynamic adaptation module that balances shared knowledge with task-specific specialization.

As shown in Figure~\ref{lora}, I-LoRA decomposes the weight update into a shared low-rank matrix $A$, a set of specialized LoRA heads $\{B_i\}_{i=1}^n$, and an interaction-aware router $R$. This design is based on the insight that diverse audio-visual tasks share a common perceptual foundation. We thus explicitly encode this general capability within the shared matrix $A$, ensuring parameter efficiency and knowledge sharing. To accommodate the divergent interaction patterns required by different tasks, the router $R$ and multiple $B$ heads operate cooperatively to decouple task-specific adaptation from shared representations. Unlike rigid task-specific assignment, I-LoRA employs a token-level soft routing strategy. It dynamically synthesizes weights to meet the semantic requirements of each token, minimizing parameter interference within the unified model.\looseness=-1

Formally, the input $H$ is constructed by concatenating the visual tokens $H_v \in \mathbb{R}^{L_v \times h}$, audio tokens $H_a \in \mathbb{R}^{L_a \times h}$, and text tokens $H_t \in \mathbb{R}^{L_t \times h}$ along the sequence~dimension:\looseness=-1
\begin{equation}
H = [\, H_v ; H_a ; H_t \,] \in \mathbb{R}^{L \times h},
\end{equation}
where $L = L_v + L_a + L_t$ is the total sequence length. Within the LLM backbone, I-LoRA is applied to the frozen linear transformation $W_0 \in \mathbb{R}^{d \times h}$. The architecture employs a shared low-rank matrix $A \in \mathbb{R}^{r \times h}$ for the initial down-projection. The interaction-aware router, parameterized by $W_r \in \mathbb{R}^{n \times r}$, utilizes the projected features $H A^\top$ to compute routing scores:
\begin{equation}
S = \mathrm{softmax}((H A^\top) W_r^\top) \in \mathbb{R}^{L \times n},
\end{equation}
where $S_{t,i}$ represents the routing weight of the $t$-th token for the $i$-th head.
The update $\Delta H$ aggregates outputs from $n$ heads $\{B_i \in \mathbb{R}^{d \times r}\}_{i=1}^n$, weighted by their respective routing scores:
\begin{equation}
\Delta H = \sum_{i=1}^{n} \mathrm{diag}(S_{:,i}) (H A^\top B_i^\top),
\end{equation}
where $\mathrm{diag}(S_{:,i}) \in \mathbb{R}^{L \times L}$ applies token-wise gating. Finally, the output of the I-LoRA-augmented layer is:
\begin{equation}
H' = H W_0^\top + \Delta H,
\end{equation}
where $H' \in \mathbb{R}^{L \times d}$ is the updated hidden representation.

We present a unified interface that standardizes tasks across three AV-LLM paradigms into a single-stage pipeline, trained under the standard next-token prediction objective of LLMs. With the I-LoRA module, we manage the trade-off between shared and specific parameters effectively. This method reconciles the conflicts caused by task heterogeneity, unlocking synergy for unified learning in complex environments.
\section{Experiments}

This section presents a comprehensive evaluation of our method. We begin by defining the experiment setup, including the tasks, datasets, and metrics. We then detail the training strategies for various AV-LLM architectures alongside their implementation. Our analysis starts by benchmarking the approach against representative baselines, covering both unified AV-LLMs and specialized non-LLM models. Following this, we compare single-task versus multi-task learning and visualize I-LoRA to substantiate our design rationale. We also conduct extensive ablation studies to investigate the effectiveness of I-LoRA, the contribution of heads through dropping, and the sensitivity to the number of $B$-heads. We conclude with qualitative case studies to demonstrate the effectiveness of our method in handling complex audio-visual contexts.\looseness=-1

\subsection{Tasks, Datasets and Evaluation Metrics}
\label{metrics}

\begin{table*}[t]
\centering
\caption{Performance comparison with unified audio-visual LLMs.}
\label{tab:av_comparison_merged}
\footnotesize
\renewcommand{\arraystretch}{1}
\setlength{\tabcolsep}{1pt}

\sbox{\tablebox}{%
\begin{tabular}{l ccc cc cc cc cc cc c c cc cc cc cc c c cc}
\toprule
\multirow{3}{*}{\textbf{Method}} & \multirow{3}{*}{\makecell{\textbf{MUSIC}\\\textbf{ACC}}} & \multirow{3}{*}{\makecell{\textbf{AVQA}\\\textbf{ACC}}} & \multirow{3}{*}{\makecell{\textbf{AVE}\\\textbf{ACC}}} & \multicolumn{2}{c}{\textbf{AVVP}} & \multicolumn{2}{c}{\textbf{ARIG}} & \multicolumn{4}{c}{\textbf{AVS}} & \multicolumn{4}{c}{\textbf{Ref-AVS}} & \multirow{3}{*}{\makecell{\textbf{CRE}\\\textbf{ACC}}} & \multicolumn{2}{c}{\textbf{DFEW}} & \multicolumn{2}{c}{\textbf{MAFW}} & \multicolumn{2}{c}{\textbf{MELD}} & \multirow{3}{*}{\makecell{\textbf{KS}\\\textbf{ACC}}} & \multirow{3}{*}{\makecell{\textbf{UCF51}\\\textbf{ACC}}} & \multicolumn{2}{c}{\textbf{VGG-CM}} \\
\cmidrule(lr){5-6} \cmidrule(lr){7-8} \cmidrule(lr){9-12} \cmidrule(lr){13-16} \cmidrule(lr){18-19} \cmidrule(lr){20-21} \cmidrule(lr){22-23} \cmidrule(lr){26-27}
& & & & \multirow{2}{*}{\makecell{\textbf{Segment-}\\\textbf{level}}} & \multirow{2}{*}{\makecell{\textbf{Event-}\\\textbf{level}}} & \multirow{2}{*}{\textbf{IoU}} & \multirow{2}{*}{\textbf{AUC}} & \multicolumn{2}{c}{\textbf{S4}} & \multicolumn{2}{c}{\textbf{MS3}} & \multicolumn{2}{c}{\textbf{Seen}} & \multicolumn{2}{c}{\textbf{Unseen}} & & \multirow{2}{*}{\textbf{WAR}} & \multirow{2}{*}{\textbf{UAR}} & \multirow{2}{*}{\textbf{WAR}} & \multirow{2}{*}{\textbf{UAR}} & \multirow{2}{*}{\textbf{ACC}} & \multirow{2}{*}{\textbf{WAF}} & & & \multirow{2}{*}{\makecell{\textbf{A$\rightarrow$V}\\\textbf{ACC}}} & \multirow{2}{*}{\makecell{\textbf{V$\rightarrow$A}\\\textbf{ACC}}} \\
\cmidrule(lr){9-10} \cmidrule(lr){11-12} \cmidrule(lr){13-14} \cmidrule(lr){15-16}
& & & & & & & & \textbf{$\mathcal{J}$} & \textbf{$\mathcal{F}$} & \textbf{$\mathcal{J}$} & \textbf{$\mathcal{F}$} & \textbf{$\mathcal{J}$} & \textbf{$\mathcal{F}$} & \textbf{$\mathcal{J}$} & \textbf{$\mathcal{F}$} & & & & & & & & & & & & \\
\midrule
TimeChat~\cite{ren2024timechat} & -- & -- & -- & 51.28 & -- & --  & -- & -- & -- & -- & -- & -- & -- & -- & -- & -- & -- & -- & -- & -- & -- & -- & -- & -- & -- & -- \\
MEERKAT~\cite{chowdhury2024meerkat} & -- & -- & -- & 54.96 & -- & -- &  -- & -- & -- & -- & -- & -- & -- & -- & -- & -- & -- & -- & -- & -- & -- & -- & -- & -- & -- & -- \\
GroundingGPT~\cite{li2024groundinggpt} & --& -- & -- & -- & -- & \underline{44.02} & \underline{45.00} & -- & -- & -- & -- & -- & -- & -- & -- & -- & -- & -- & -- & -- & -- & -- & -- & -- & -- & -- \\
AnyRef~\cite{he2024multi} & --& -- & -- & -- & -- & -- & -- & \textbf{82.80} & \textbf{90.80} & \underline{55.60} & \textbf{66.30}  & -- & -- & -- & -- & -- & -- & -- & -- & -- & -- & -- & -- & -- & -- & -- \\
SimToken~\cite{jin2025simtoken} & --& -- & -- & -- & -- & -- & -- &-- & -- & -- & -- & \textbf{72.00} & \textbf{81.30} & \textbf{69.80} & \textbf{79.10}  & -- & -- & -- & -- & -- & -- & -- & -- & -- & -- & -- \\
PandaGPT~\cite{su2023pandagpt}& 33.70& 79.80 & -- & -- & -- & -- & -- & --  & -- & -- & -- & -- & -- & -- & -- & -- & -- & -- & -- & -- & -- & -- & -- & -- & -- & -- \\
MacawLLM~\cite{lyu2023macaw} & 31.80& 78.70 & -- & -- & -- & -- & -- & --  & -- & -- & -- & -- & -- & -- & -- & -- & -- & -- & -- & -- & -- & -- & -- & -- & -- & -- \\
Video LLaMA~\cite{zhang2023video} & 36.60& 81.00& -- & -- & -- & -- & -- &  -- & -- & -- & -- & -- & -- & -- & -- & -- & 35.75$^{\dagger}$ & 26.77$^{\dagger}$ & -- & -- & -- & -- & -- & -- & -- & -- \\
Video LLaMA 2~\cite{cheng2024videollama} & 73.60& -- & -- & -- & --  & -- & -- & -- & -- & -- & -- & -- & -- & -- & -- & -- & -- & -- & -- & -- & -- & -- & -- & -- & -- & -- \\
CAT+~\cite{ye2025cat} & \textbf{86.40}& \textbf{92.60} & -- & --  & -- & -- & -- & -- & -- & -- & -- & -- & -- & -- & -- & -- & -- & -- & -- & -- & -- & -- & -- & -- & -- & -- \\
OMCAT~\cite{goel2024omcat} & 73.80& 90.20& -- & -- & -- & -- & -- & -- & -- & -- & -- & -- & -- & -- & -- & -- & -- & -- & -- & -- & -- & -- & -- & -- & -- & -- \\
UIO-2~\cite{lu2024unified} &-- &-- & -- & -- & -- & -- & -- & -- & -- & -- & -- & -- & -- & -- & --  & -- & -- & -- & -- & -- & -- & -- & \underline{89.30} & -- & -- & -- \\
VITA~\cite{fu2024vita} & -- & -- & -- & -- & -- & -- & -- & -- & --  & -- & -- & -- & -- & -- & -- & -- & 32.07$^{\dagger}$& 21.36$^{\dagger}$ & \underline{33.38$^{\dagger}$} & \underline{14.05$^{\dagger}$} & -- & -- & -- & -- & -- & -- \\
Emotion-LLaMA~\cite{cheng2024emotion} & --& -- & -- & -- & -- & -- & -- & --  & -- & -- & -- & -- & -- & -- & -- & -- & \textbf{77.06$^{\dagger}$} & \textbf{64.21$^{\dagger}$}& -- & -- & -- & -- & -- & -- & -- & -- \\
\midrule
Crab~\cite{du2025crab} & 78.94& -- & \underline{80.15}& \underline{59.00}& \underline{54.44}& 41.78& 42.00& 73.25& 86.81& \textbf{58.21}& \underline{66.24}& 40.54& 58.00& 45.55& 63.00& --&-- & --& --& --& --& --& --& --& --& --\\
\rowcolor[gray]{0.95}
Crab$^{+}$ & \underline{81.09}& \underline{92.16}& \textbf{83.58}& \textbf{59.47}& \textbf{55.79}& \textbf{79.62}& \textbf{79.60}& \underline{75.75}& \underline{86.94}& 52.64& 64.27& \underline{65.45}& \underline{72.68}& \underline{68.98}& \underline{74.41}& \textbf{84.41}& \underline{64.34}& \underline{55.83}& \textbf{45.60}& \textbf{32.40}& \textbf{52.08}& \textbf{51.14}& \textbf{91.12}& \textbf{94.04}& \textbf{74.50}& \textbf{76.97}\\
\bottomrule
\end{tabular}
}

\newdimen\scaledwidth
\scaledwidth=\dimexpr 0.62\wd\tablebox\relax

\scalebox{0.62}{\usebox{\tablebox}}

\vspace{1em}
\begin{minipage}{\scaledwidth}
\scriptsize
\textit{Note:} The symbol -- denotes that the task is not applicable or results are not reported. The symbol $^\dagger$ identifies datasets containing multiple standard splits; we exclusively report results on the first split to accommodate the computational complexity of multi-task learning.
\end{minipage}
\end{table*}

\textbf{Audio-visual action recognition.}
This task aims to recognize human actions from audio-visual streams. Evaluations are performed on Kinetics-Sounds (KS)\cite{arandjelovic2017look}, covering 31 action classes, and UCF51, a 51-class subset of UCF101\cite{soomro2012ucf101}. We adopt the official dataset split for KS and the first predefined split for UCF51, reporting Top-1~accuracy.\looseness=-1

\textbf{Audio-visual emotion recognition.}
This task predicts expressed emotions from audio-visual inputs. We evaluate on CREMA-D~\cite{cao2014crema}, MAFW~\cite{liu2022mafw}, DFEW~\cite{jiang2020dfew}, MER2024~\cite{lian2024mer}, and MELD~\cite{poria2018meld}. CREMA-D contains acted emotional speech, while the other datasets consist of natural expressions from TV or film. We adhere to the split protocol of~\cite{peng2022balanced,huang2025adaptive,wei2024fly} for CREMA-D, the first official split for MAFW and DFEW, and the official division for MELD. MER2024 is trained on its ``Train\&Val'' set. Following established benchmarks, we report accuracy for CREMA-D, accuracy and Weighted F1 (WAF) for MELD, and UAR/WAR for the remaining datasets.

\textbf{Cross-modal matching.}
This binary classification task determines whether an audio clip and a video correspond to the same event. We evaluate our proposed VGGSound Cross-modal Matching (VGG-CM) benchmark and report~accuracy.\looseness=-1

\textbf{Audio-visual event localization and parsing.}
Event localization determines the temporal boundaries of audio-visual events, while parsing classifies these events as either unimodal or multimodal. We adopt AVE~\cite{tian2018audio} and UnAV-100~\cite{geng2023dense} for localization, and LLP~\cite{tian2020unified} for parsing.
The AVE dataset contains 4,143 10-second clips across 28 categories, whereas UnAV-100 covers 30,059 annotated events from 100 categories within untrimmed videos. LLP comprises 11,849 curated YouTube videos. Following standard protocols, we report localization accuracy on AVE and F1 scores at both the segment and event levels (IoU 0.5) on LLP. To mitigate the computational cost of UnAV-100's long sequences, we train on a sampled subset of clips with stable temporal dynamics, thus maintaining event coherence even with reduced input data.\looseness=-1

\textbf{Audio-visual localization.}
This task aims to segment and localize sounding objects. We use AVSBench~\cite{zhou2022audio}, which provides pixel-level masks for Single-Source (S4) and Multi-Source (MS3) settings. To evaluate reference grounding, we also include Ref-AVS with its Seen and Unseen subsets. Consistent with prior work, we take the first frame for S4 and uniformly sampled frames for MS3. Additionally, the S4 bounding boxes serve as annotations for Audio-Referred Image Grounding (ARIG). For Ref-AVS, representative frames are sampled from each 10-second clip. We report performance using the Jaccard index $\mathcal{J}$ and F-score $\mathcal{F}$. For ARIG, we adopt Intersection-over-Union (IoU) and Area Under the Curve (AUC) as evaluation metrics.

\textbf{Audio-visual captioning.}
This task generates natural-language descriptions that reflect audio-visual semantics. We use the subset of VALOR~\cite{chen2023valor} to strengthen the model’s fundamental audio-visual perceptual capability.

\textbf{Audio-visual question answering.} 
This task entails inferring answers using joint acoustic and visual cues. We evaluate our method on MUSIC-AVQA~\cite{li2022learning}, which comprises musical scenes spanning 22 instrument categories, and AVQA~\cite{yang2022avqa}, containing 57.3K QA pairs across 9 question types. We report accuracy based on the official splits.\looseness=-1

\subsection{Training Strategy}

We tailor the training process to the specific architecture of each model. Since LLM+V+A and V-LLM+A lack inherent audio-visual priors, we train them in two stages: (1) modality-specific connector pre-training, followed by (2) audio-visual multi-task instruction tuning. The native AV-LLM, which already has basic audio-visual perception, proceeds directly to single-stage instruction tuning.

\textbf{Stage 1: Modality-specific connector pre-training}.
For LLM+V+A, the visual connector is pre-trained on a dataset from VideoLLaVA~\cite{lin2023video}, with all other parameters frozen, thereby establishing basic visual perception.
For audio processing in both LLM+V+A and V-LLM+A, the audio connector is pre-trained on AudioSet~\cite{gemmeke2017audio}. Due to AudioSet’s coarse labels, we employ Qwen2Audio~\cite{chu2024qwen2} to generate captions as supervision. Only audio connector parameters are updated.

\textbf{Stage 2: Audio-visual multi-task instruction tuning}.
All three paradigms are fine-tuned on the AV-UIE v2 dataset. By default, only interaction-aware LoRA is optimized. For LLM+V+A, we additionally fine-tune audio and visual connectors to improve perceptual alignment. For V-LLM+A and native AV-LLM, connectors are frozen to maintain the cross-modal consistency of their pre-trained~encoders.\looseness=-1

\subsection{Implementation Details}

In the audio-visual emotion recognition task, we sample four frames per video, given the short clip duration, whereas ten frames are used for all other tasks. All visual inputs are resized to \(224\times224\). Raw audio waveforms are resampled to 16 kHz. We set the number of learnable tokens \(K\) to 32 for both visual and audio Q-Formers. For the window-level Q-Former, each 0.33-second audio segment maps to a single learnable token, resulting in 88 tokens for a 30-second clip. The segmentation model is initialized with pre-trained SAM2.1-Large weights~\cite{ravi2024sam}. We adjust the hyperparameters of the interaction-aware LoRA modules based on the experiment setup. To maximize the capacity of the native AV-LLM in Crab$^{+}$, we set the rank \(r=128\), scaling factor \(\alpha=256\), and dropout rate to 0.10. For consistency with the analytical experiments in Section~\ref{intro} (validating multitask effectiveness), we use $r=16$, $\alpha=32$, and a dropout rate of 0.05. The LoRA $B$ matrix employs 3 heads.

\subsection{Results and Analysis}

\begin{table}[t!]
\centering
\caption{Comparison of methods on recognition tasks.}
\label{tab:expert4}
\small
\renewcommand{\arraystretch}{0.8}
\setlength{\tabcolsep}{4pt}
\scalebox{0.65}{
\begin{tabular}{l c c c c c c}
\toprule
\textbf{Method} & \textbf{CREMA-D} & \textbf{MELD} & \textbf{DFEW} & \textbf{MAFW} & \textbf{KS} & \textbf{UCF51}\\
\midrule
OGM-GE~\cite{peng2022balanced} & 64.34 & -- & -- & --  & 66.35 & -- \\
MMPareto~\cite{wei2024mmpareto} & 70.19 & -- &  -- & --& 69.13 & -- \\
CAL~\cite{xu2025contribution} & \underline{79.30} & -- & --& --& 74.82 & -- \\
DialogueRNN~\cite{majumder2019dialoguernn} & -- & 56.12 & -- & --& --& -- \\
DQ-Former~\cite{jing2024dq} & --  & \underline{64.88} & --& --& --& --\\
ACoSDN~\cite{ding2025disentangling} & -- & \textbf{68.04} & --& --& --& -- \\
DFER-CLIP~\cite{zhao2023prompting} & -- & -- & 71.25$^{\dagger}$ & 52.59$^{\dagger}$& --& -- \\
VAEmo~\cite{cheng2025vaemo} & -- & -- & \underline{75.78$^{\dagger}$} & \underline{58.91$^{\dagger}$}& --& -- \\
AVF-MAE++~\cite{wu2025scalable} & -- & -- & \textbf{77.45$^{\dagger}$} & \textbf{60.24$^{\dagger}$}& --& -- \\
AttenClus~\cite{long2018attention} & -- & -- & -- & --& 73.91& 84.79$^{\dagger}$ \\
DMRN~\cite{tian2018audio} & -- & -- & -- & --& 77.50& 82.93$^{\dagger}$ \\
MAFnet~\cite{brousmiche2022multimodal} & -- & -- & -- & --& \underline{83.94}& \underline{86.72$^{\dagger}$} \\
\midrule
Crab$^{+}$ & \textbf{84.41} &52.08 &64.34 & 45.60 & \textbf{91.12} & \textbf{94.04}\\
\bottomrule
\multicolumn{7}{l}{%
  \begin{tabular}[c]{@{}l@{}}
  \vspace{1pt} \\ 
  \footnotesize \textit{Note:} $^{\dagger}$ denotes results averaged over all splits. Due to the computational complexity of multi-task \\
  \footnotesize learning, we use the first split for evaluation.
  \end{tabular}%
}
\end{tabular}
}
\end{table}

\textbf{Comparison with Audio-Visual Unified Models.} Table~\ref{tab:av_comparison_merged} presents a comparative analysis of our approach against existing unified AV-LLMs. In contrast to prior models restricted to limited task subsets, our model encompasses a wider array of audio-visual scenarios. Notably, Crab$^{+}$ represents a substantial advancement over Crab, exhibiting a broader spectrum of task execution capabilities while achieving superior performance across the majority of benchmarks. Compared to other unified AV-LLMs, our method achieves superior performance on tasks such as AVE (83.58 vs. 80.15), AVVP (59.47 vs. 59.00), and KS (91.12 vs. 89.30). Furthermore, it yields competitive results on tasks including AVS, Ref-AVS, and DFEW. This demonstrates that our proposed method effectively balances task diversity with specific proficiency.

\begin{table}[t!]
\centering
\caption{Comparison of methods on temporal localization tasks.}
\label{tab:expert3}
\small
\renewcommand{\arraystretch}{0.8}
\setlength{\tabcolsep}{4pt}
\scalebox{0.90}{
\begin{tabular}{l cc c}
\toprule
\multirow{2}{*}{\textbf{Method}}
& \multicolumn{2}{c}{\textbf{AVVP}}
& \multirow{2}{*}{\textbf{AVE}} \\
\cmidrule(lr){2-3}
& \textbf{Segment-level} & \textbf{Event-level} & \\
\midrule
AVT~\cite{lin2020audiovisual} & -- & --&75.80 \\
MM-Pyramid~\cite{yu2022mm} & 59.20 & 53.04 & 77.80 \\
CMBS~\cite{xia2022cross} & 55.00& 48.48& -- \\
DHHN~\cite{jiang2022dhhn} &\textbf{60.32}& \underline{55.06}& -- \\
MoLT~\cite{rho2025molt} &--& --& \underline{83.50} \\
\midrule
Crab$^{+}$ & \underline{59.47}& \textbf{55.79} &\textbf{83.58} \\
\bottomrule
\end{tabular}
}
\end{table}

\begin{table}[t!]
\centering
\caption{Comparison of methods on spatial localization tasks.}
\label{tab:expert2}
\small
\renewcommand{\arraystretch}{0.8}
\setlength{\tabcolsep}{4pt}
\scalebox{0.85}{
\begin{tabular}{l c c c c c}
\toprule
\textbf{Method} & \textbf{ARIG} & \textbf{S4} & \textbf{MS3} & \textbf{Seen} & \textbf{Unseen}\\
\midrule
LVS~\cite{chen2021localizing} & 23.69 & -- & -- & -- &--\\
FNAC~\cite{sun2023learning} & 27.15 & -- & -- & -- & -- \\
AVSBench~\cite{zhou2022audio} & -- & \underline{78.70} & \underline{54.00} & 23.20 & 32.36\\
AVSegFormer~\cite{gao2024avsegformer} & -- & \textbf{83.06} & \textbf{61.33} & 33.47 & 36.05\\
EEMC~\cite{wang2024ref} & -- & -- & -- & 34.20 & 49.54\\
SAM2-LOVE~\cite{wang2025sam2} & -- & -- & -- & \underline{43.50} & \underline{66.50}\\
\midrule
Crab$^{+}$ & \textbf{79.62} & 75.75 & 52.64 & \textbf{65.45} & \textbf{68.98}\\
\bottomrule
\end{tabular}
}
\end{table}

\begin{table}[t!]
\centering
\caption{Comparison of methods on AVQA tasks.}
\label{tab:expert1}
\small
\renewcommand{\arraystretch}{0.8}
\setlength{\tabcolsep}{4pt}
\scalebox{0.80}{
\begin{tabular}{l cccc c}
\toprule
\multirow{2}{*}{\textbf{Method}}
& \multicolumn{4}{c}{\textbf{MUSIC-AVQA}}
& \multirow{2}{*}{\textbf{AVQA}} \\
\cmidrule(lr){2-5}
& \textbf{Audio} & \textbf{Visual} & \textbf{Audio-Visual} & \textbf{Overall} & \\
\midrule
ST-AVQA~\cite{li2022learning} & 73.87 &74.40 &69.53 & 71.59 & -- \\
PSTP-Net~\cite{li2023progressive} & 70.91 & 77.26 & 72.57 & 73.52 & \underline{90.20} \\
SaSR-Net~\cite{yang2024sasr} & 73.56 & 73.28 & \underline{74.66} & 74.21 & 89.90 \\
QA-TIGER~\cite{kim2025question} & \underline{78.58}& \underline{85.14}& 73.74& \underline{77.62}& -- \\
HCRN+HAVF\cite{le2020hierarchical} & --& -- &-- & -- & 89.00 \\
\midrule
Crab$^{+}$ & \textbf{79.44}& \textbf{92.45} &\textbf{76.20} & \textbf{81.09} & \textbf{92.16}\\
\bottomrule
\end{tabular}
}
\end{table}

\begin{figure*}[!t]
\centering
\includegraphics[width=1\textwidth]{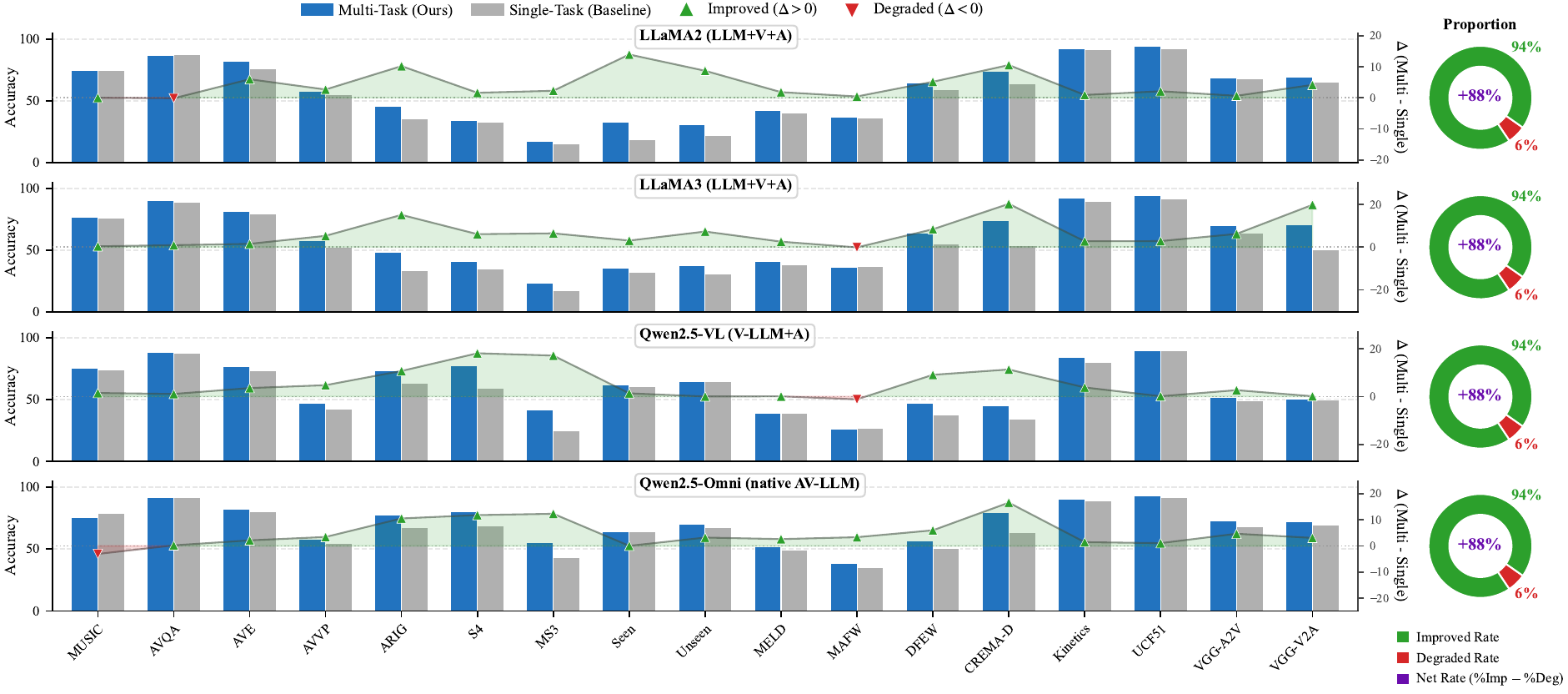}
\caption{Comparative analysis of synergy in multi-task vs. single-task learning across three representative AV-LLM paradigms. For each model, we report the proportion of tasks exhibiting positive gains and the net score, defined as the difference between the percentage of tasks benefiting from multi-task learning and those suffering negative transfer. The results demonstrate that our method effectively mitigates task heterogeneity, promoting inter-task synergy.}
\label{synergy}
\end{figure*}

\textbf{Comparison with Specialized Audio-Visual Models.}
Although our unified model is built for general audio-visual scene understanding, it remains highly effective against task-specific expert models. Regarding the recognition tasks reported in Table~\ref{tab:expert4}, the results underscore the inherent difficulty of multi-task integration. While a minor performance gap remains on several emotion recognition benchmarks, our model demonstrates solid performance on CREMA-D with a score of 84.41 and achieves superior results in action recognition, specifically 91.12 on KS and 94.04 on UCF51. For temporal localization, as shown in Table~\ref{tab:expert3}, our method remains highly competitive. It achieves an accuracy of 83.58 on the AVE dataset, slightly outperforming the previous best-performing approach. On the AVVP task, the model attains the highest event-level score of 55.79, while simultaneously maintaining comparable performance at the segment-level. The advantages of our model are most pronounced in spatial grounding tasks. Table~\ref{tab:expert2} shows that our method achieves substantial improvements over dedicated baselines, surpassing prior methods by approximately +52.47 on ARIG and +21.95 on the Ref-AVS (Seen) subset. In addition, Crab$^{+}$ demonstrates strong performance on question answering tasks, as reported in Table~\ref{tab:expert1}. Benefiting from robust visual perception, it attains an overall accuracy of 81.09 on MUSIC-AVQA and 92.16 on the AVQA benchmark. Overall, these results suggest that the proposed unified model achieves performance comparable to task-specific methods.\looseness=-1

\textbf{Single-Task vs. Multi-Task.}
After validating the performance of Crab$^{+}$, we further investigate the generalizability of our unified learning method across distinct AV-LLM construction paradigms (e.g., LLM+V+A and V-LLM+A). We conduct a comparative analysis of single-task and multi-task settings on the AV-UIE v2 dataset using I-LoRA. To ensure a fair comparison, we maintain consistent training configurations, varying only the data composition: single-task models are trained on individual datasets, while the multi-task model is trained on the comprehensive dataset encompassing all tasks. As shown in Figure~\ref{synergy}, under our unified learning method, AV-LLMs across different paradigms consistently outperform single-task baselines in the multi-task setting. Quantitatively, our approach achieves performance improvements in 94\% of tasks, with only minor negative transfer observed in remaining tasks, resulting in a substantial net gain of 88\%. This contrasts with the analysis in Section \ref{intro}, where naïve multi-tasking (utilizing isolated instruction following with LoRA) causes performance degradation in approximately 55\% of tasks, leading to a negative net gain of -10\%. By increasing the positive gain rate from 45\% to 94\% and improving the net gain from -10\% to +88\%, our results demonstrate that our method effectively mitigates task heterogeneity, thereby facilitating robust positive transfer and synergy among tasks.

\begin{figure}[!htbp]
\centering
\includegraphics[width=0.45\textwidth]{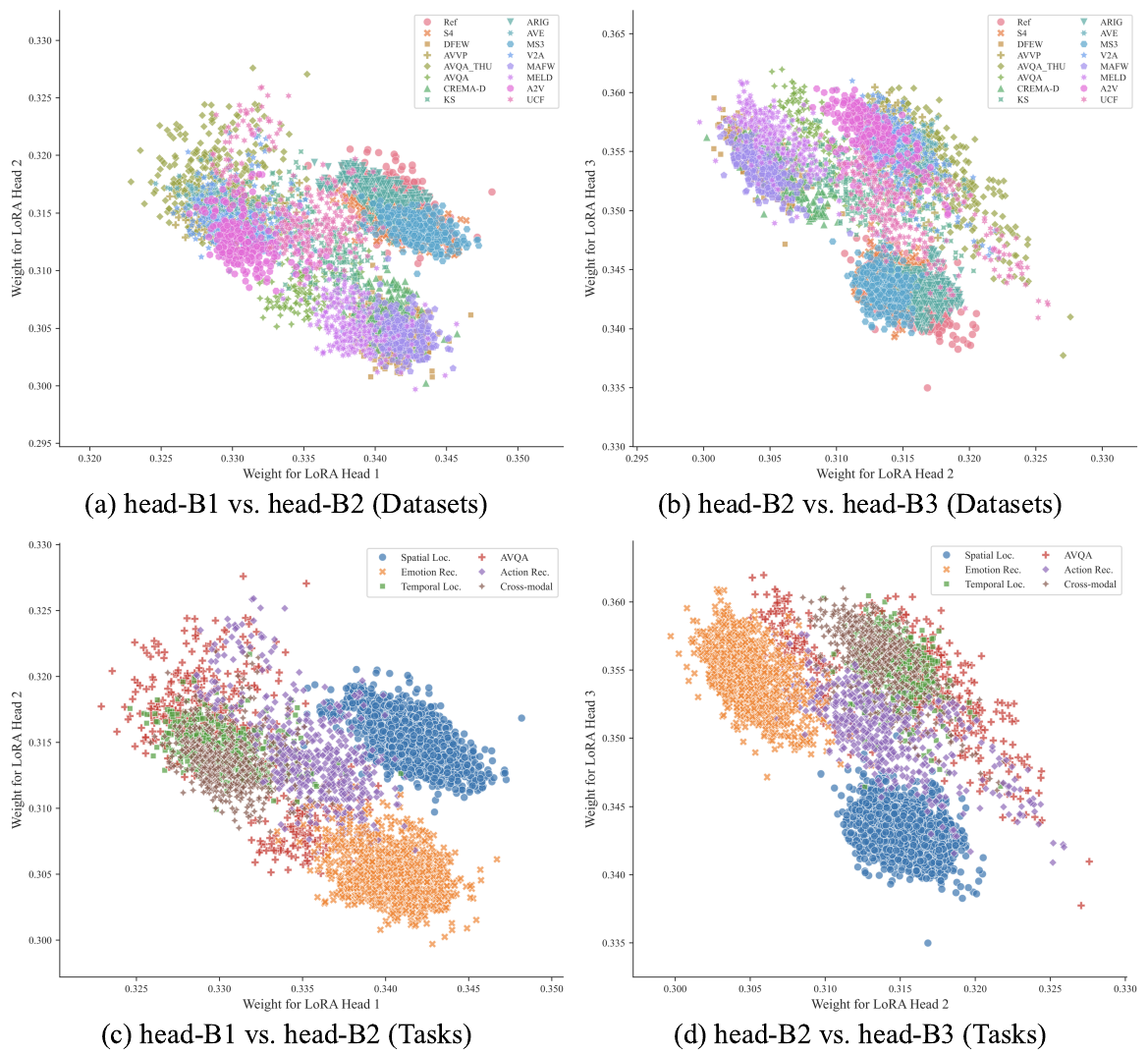}
\caption{Visualization of router weights. Head pair comparisons across (a)–(b) datasets and (c)–(d) tasks. Tasks sharing similar demands form distinct clusters, indicating capacity to clarify complex audio-visual patterns.}
\label{router}
\end{figure}

\begin{figure}[!htbp]
\centering
\includegraphics[width=0.50\textwidth]{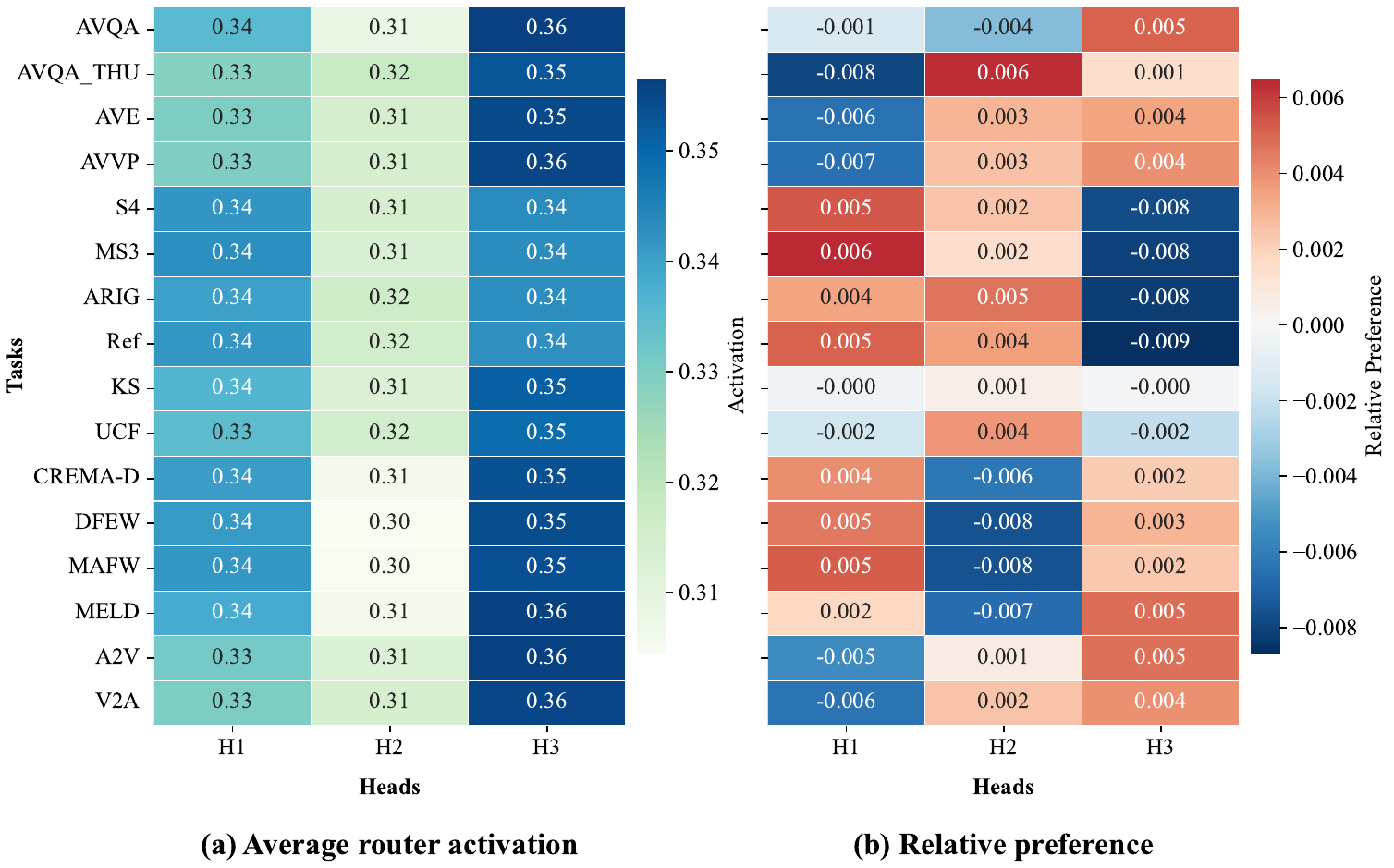}
\caption{Router activation analysis. (a) Average activations show balanced contributions ($\approx 0.3$). (b) Relative patterns reveal task-specific head preferences.}
\label{router_v2}
\end{figure}

\begin{figure}[!htbp]
\centering
\includegraphics[width=0.45\textwidth]{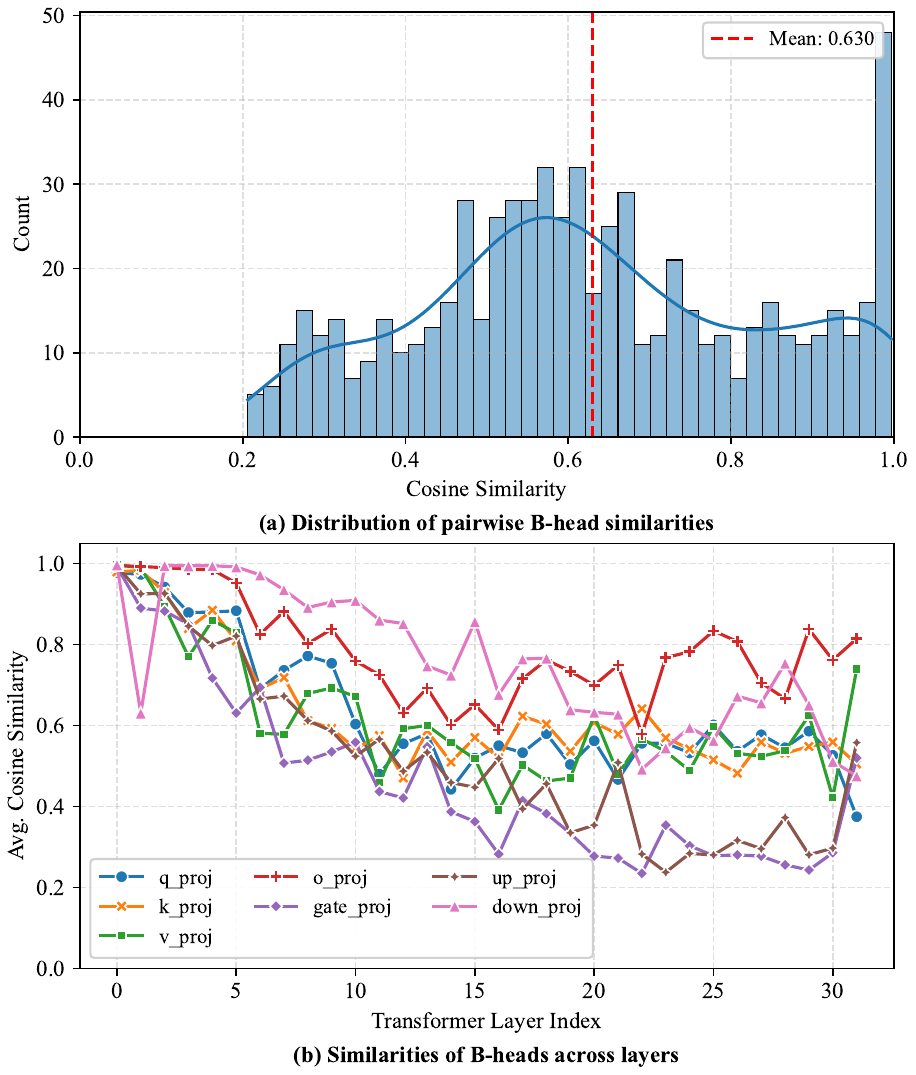}
\caption{Similarity analysis of routing matrices $B$. (a) Overall similarity (mean 0.630). (b) Layer-wise similarity, showing divergence in deeper layers.}
\label{router_v3}
\end{figure}

\textbf{Visualization Analysis}.
To examine the internal adaptability of I-LoRA, we analyze router weights, activation distributions, and parameter similarities (Figures \ref{router}–\ref{router_v3}). Although I-LoRA adopts token-level soft routing rather than task-level hard routing, the router weights exhibit clear semantic clustering. Clusters emerge not only across datasets but also across tasks, such as spatial localization (e.g., ARIG and Ref-AVS), with datasets from the same task forming compact groups. This suggests that the router effectively captures task-specific semantics. We further analyze head-wise activation behaviors. Overall, all heads maintain a relatively balanced activation across diverse inputs, with an average activation probability of approximately 0.3. Despite this balance, different heads exhibit distinct functional preferences: head-B1 focuses on spatial localization and emotion recognition, head-B2 emphasizes temporal localization, and head-B3 specializes in AVQA and cross-modal matching. To validate this functional differentiation, we compute pairwise similarities between flattened weight vectors of the three heads, averaged across all layers and LoRA modules. The mean similarity of 0.630 suggests partial parameter sharing alongside role specialization. Layer-wise analysis shows higher similarity in shallow layers and progressively lower similarity in deeper layers, indicating that task-specific differentiation becomes more pronounced with increasing network depth.

\begin{table}[bhp!]
\centering
\caption{Results of dropping heads.}
\label{tab:drophead}
\small
\renewcommand{\arraystretch}{1}
\setlength{\tabcolsep}{4pt}
\scalebox{0.70}{
\begin{tabular}{l| c c c c c c c c}
\toprule
\textbf{Method} & \textbf{CREMA-D} & \textbf{AVQA} & \textbf{MUSIC} & \textbf{AVVP} & \textbf{KS} & \textbf{ARIG} & \textbf{A$\rightarrow$V} & \textbf{V$\rightarrow$A}\\
\midrule
drop head-1 & 74.06 &86.57 &71.11 & 51.91 &91.01 &40.22&68.14&66.15 \\
drop head-2 & 72.37  &85.30& 71.55& 54.65 & 91.57&43.94&67.01&67.83\\
drop head-3& 71.76  &84.34&69.80 & 39.72 &90.76 & 44.40&67.88&67.14\\
\textbf{no drop} & \textbf{74.12} & \textbf{86.92}& \textbf{74.73}& \textbf{59.77} &\textbf{92.21} &\textbf{45.78} &\textbf{68.81}&\textbf{69.28} \\
\bottomrule
\end{tabular}
}
\end{table}

\begin{figure}[!t] \centering \includegraphics[width=0.48\textwidth]{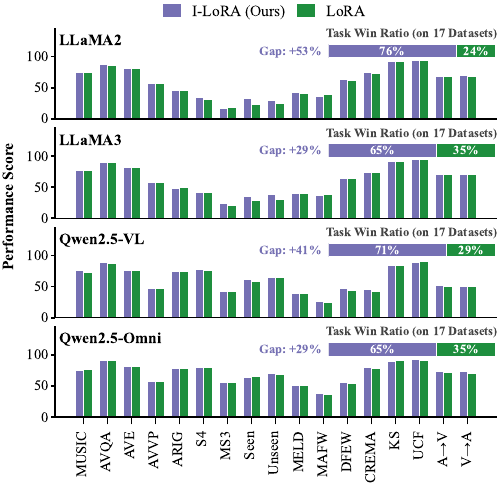} \caption{Ablation results about LoRA. While LoRA shows gains on a subset of tasks, I-LoRA significantly increases the number of tasks with synergy.} \label{fig:lora_ablation} \end{figure}

\begin{figure}[t!] \centering \includegraphics[width=0.48\textwidth]{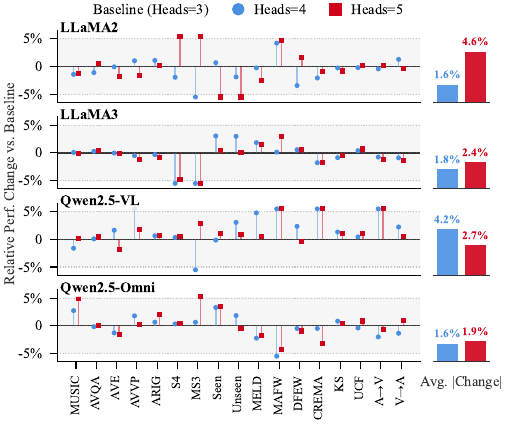} \caption{Ablation on the number of $B$ heads. Our method remains stable without significant fluctuations across this~hyperparameter.\looseness=-1} \label{fig:heads_ablation} \end{figure}

\begin{figure*}[t!]
\centering 
\includegraphics[width=0.95\textwidth]{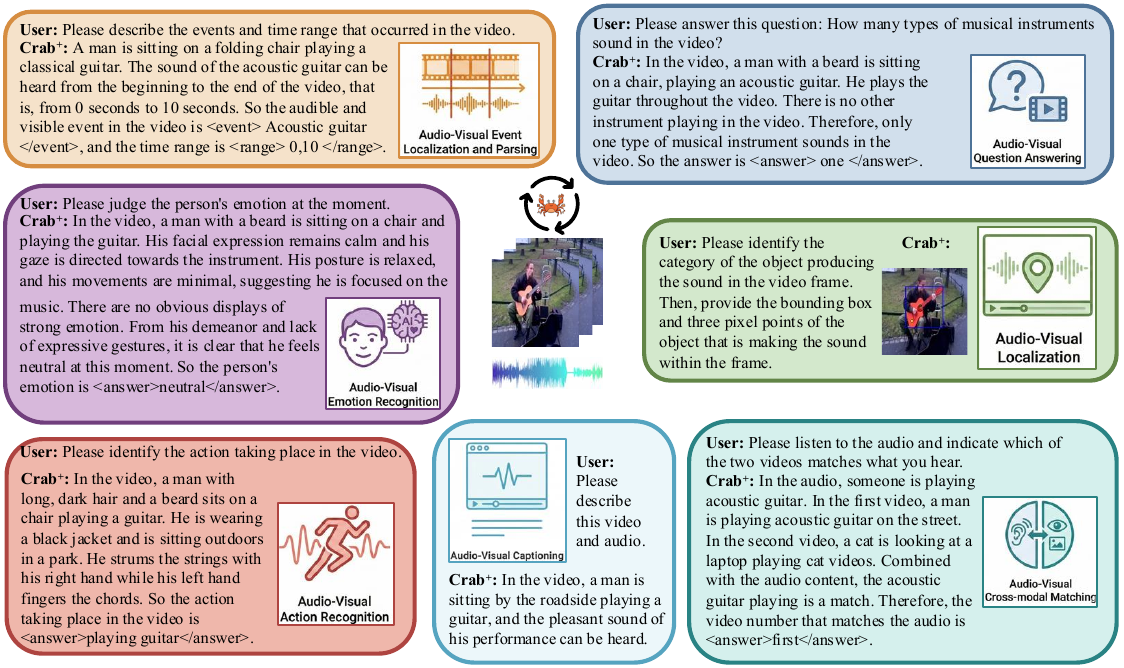} 
\caption{Qualitative results. Crab$^{+}$ is capable of performing multiple different audio-visual scene understanding tasks on a single~sample.\looseness=-1} 
\label{fig:qualitative} 
\end{figure*}

\textbf{Ablation Study on LoRA.} To verify the efficacy of I-LoRA, we benchmark it against the LoRA baseline. We adopt a unified multi-task training method on the AV-UIE v2 dataset, aiming to mitigate the inherent task heterogeneity in the data. As shown in Figure~\ref{fig:lora_ablation}, the results indicate that joint training across tasks can already yield performance gains, as evidenced by LoRA achieving better results on a limited subset of tasks. This observation reflects the inherent potential of multi-task learning. By explicitly addressing parameter interference through inter-task interaction, I-LoRA further amplifies and stabilizes such positive transfer, leading to superior performance on the majority of benchmarks. Across different model architectures, our method achieves the best performance in 65\%–76\% of the tasks, representing a 29\%–53\% absolute improvement over the LoRA baseline in terms of task coverage.\looseness=-1

\textbf{Head Ablation Study.} To investigate the contribution of each LoRA head, we employed a ``head-dropping'' strategy by zeroing out the parameters of specific heads within the LLaMA2 backbone. As shown in Table~\ref{tab:drophead}, removing any single head leads to consistent performance degradation across all tasks, covering AVQA, temporal localization (AVVP), spatial localization (ARIG), and cross-modal perception ($A\rightarrow V$, $V\rightarrow A$). These results suggest that each head captures complementary information rather than highly similar or redundant representations, thereby jointly modeling the audio-visual space. In addition, the varying degrees of performance degradation across tasks are consistent with our earlier visualization analysis. The agreement between quantitative and qualitative results further supports the soundness of our design. Overall, the results demonstrate that I-LoRA is suited to handling task heterogeneity in complex audio-visual scene understanding, facilitating inter-task cooperation and leading to more effective unified learning.

\textbf{Ablation Study on the Number of $B$-heads.}
We further investigate the sensitivity of I-LoRA to the number of $B$-heads. Intuitively, the number of heads determines the breadth of the adaptation subspace. A larger number theoretically provides more capacity for new tasks. However, this configuration involves a trade-off. A minimal count limits the ability to model complex tasks. Conversely, an excessive number makes it difficult to establish connections between heads for effective unified learning. To evaluate stability, we set $N=3$ as the baseline and test the impact of increasing the head count to 4 and 5 across different AV-LLMs. We then visualize both the task-specific deviation and the total average deviation relative to this baseline. As shown in Figure \ref{fig:heads_ablation}, sensitivity varies by backbone architecture. The LLaMA 2-based model exhibits the highest deviation of 4.6\% at $N=5$. Meanwhile, Qwen2.5-Omni demonstrates the lowest deviation of 1.6\%. Despite these variations, the performance fluctuations for all AV-LLMs remain within a reasonable 5\% margin relative to the baseline. These results confirm the stability of our method.

\textbf{Qualitative Case Study.} Figure~\ref{fig:qualitative} presents a qualitative analysis of a representative clip showing a man playing an acoustic guitar by the roadside. The model demonstrates consistent performance across a range of audio-visual tasks. The analysis highlights the model's capabilities progressively. Starting with basic recognition, the model accurately identifies the action of playing the guitar and infers the performer's emotional state. Moving to fine-grained perception, it precisely localizes the temporal boundaries of the event and segments the sounding instrument within the frame. It also succeeds in cross-modal matching by identifying the corresponding video segment given the input audio, despite the presence of distractors. Finally, in complex reasoning, the model comprehends and synthesizes both audio and visual content to provide a correct answer to the query, ``How many types of musical instruments sound in the video?'' This case exemplifies the seamless integration of low-level perception with high-level reasoning, validating its potential as a unified audio-visual assistant.
\section{Discussion}

We introduce Crab$^{+}$, a scalable and unified audio-visual scene understanding model designed to mitigate the severe negative transfer observed in naïve multi-task instruction tuning. Rather than viewing task heterogeneity as an obstacle, Crab$^{+}$ reconciles these differences through explicit cooperation from both data and model perspectives. On the data side, we construct AV-UIE v2 to bridge output discrepancies via explicit reasoning processes. On the model side, we design a unified input-output interface to standardize diverse task structures and propose Interaction-aware LoRA (I-LoRA) to dynamically coordinate conflicting capability demands. By aligning these heterogeneous tasks, Crab$^{+}$ successfully reverses the trend of performance degradation, converting negative interference into positive synergy and establishing a robust baseline for generalist audio-visual scene understanding.

\textbf{Limitation}. While explicit reasoning benefits complex scenarios, applying extensive reasoning chains to simpler tasks may induce over-interpretation, potentially introducing noise that interferes with direct predictions. Thus, developing adaptive data strategies that dynamically align reasoning granularity with task complexity remains an~open challenge.

\textbf{Future Work}. Moving forward, we plan to extend our unified model to incorporate generative tasks, advancing the pursuit of general-purpose audio-visual scene~ understanding.\looseness=-1

\clearpage

\bibliographystyle{IEEEtran}
\bibliography{paper}

\clearpage

\end{document}